%% file: main.tex
\documentclass[letterpaper]{article} % DO NOT CHANGE THIS
\usepackage{aaai24}  % DO NOT CHANGE THIS
\usepackage{times}  % DO NOT CHANGE THIS
\usepackage{helvet}  % DO NOT CHANGE THIS
\usepackage{courier}  % DO NOT CHANGE THIS
\usepackage[hyphens]{url}  % DO NOT CHANGE THIS
\usepackage{graphicx} % DO NOT CHANGE THIS
\usepackage{natbib}  % DO NOT CHANGE THIS AND DO NOT ADD ANY OPTIONS TO IT
\usepackage{caption} % DO NOT CHANGE THIS AND DO NOT ADD ANY OPTIONS TO IT
\usepackage{algorithm}
\usepackage{algorithmic}

\usepackage{microtype}
\usepackage{subcaption} % we added 
\usepackage{amsmath} % we added 
\usepackage{amsfonts} % we added 
\usepackage{xcolor} % we added  

\usepackage{newfloat}
\usepackage{listings}
\DeclareCaptionStyle{ruled}{labelfont=normalfont,labelsep=colon,strut=off} % DO NOT CHANGE THIS
\floatstyle{ruled}
\newfloat{listing}{tb}{lst}{}
\floatname{listing}{Listing}
\DeclareMathOperator*{\argmax}{arg\,max}

\title{Can LLMs Fix Issues with Reasoning Models?\\Towards More Likely Models for AI Planning}
\author {
    Turgay Caglar\textsuperscript{\rm 1},
    Sirine Belhaj\textsuperscript{\rm 2},
    Tathagata Chakraborti\textsuperscript{\rm 3},
    Michael Katz\textsuperscript{\rm 3},
    Sarath Sreedharan\textsuperscript{\rm 1}
}
\affiliations {
    \textsuperscript{\rm 1}Colorado State University\\
    \textsuperscript{\rm 2}Ecole Polytechnique de Tunisie\\
    \textsuperscript{\rm 3}IBM Research\\
    tcaglar@colostate.edu
}

\begin{document}

\maketitle

\begin{abstract}
This is the first work to look at the application of large language models (LLMs) for the purpose of model space edits in automated planning tasks. To set the stage for this union, we explore two different flavors of model space problems that have been studied in the AI planning literature and explore the effect of an LLM on those tasks. We empirically demonstrate how the performance of an LLM contrasts with combinatorial search (CS) -- an approach that has been traditionally used to solve model space tasks in planning, both with the LLM in the role of a standalone model space reasoner as well as in the role of a statistical signal in concert with the CS approach as part of a two-stage process. Our experiments show promising results suggesting further forays of LLMs into the exciting world of model space reasoning for planning tasks in the future.
\end{abstract}

\section{Introduction}

AI planning or automated planning (used interchangeably) 
is the task of synthesizing the goal-directed behavior of autonomous agents.
Traditionally, the AI planning community has looked at the classical
planning problem as one of generating a plan given a model 
of the world \cite{ghallab2004automated}.
Here, ``model'' or a ``planning problem'' 
refers to a collection of constraints describing 
the current state of the world (initial state), 
the actions available to the agent along with 
the conditions under which the agent
can do those actions and the effect of doing those actions on the
environment, and a target (goal) state for the agent to achieve.
The plan is a sequence of actions that the agent can use
to transform the current state
to the desired goal state.

Typically, these models are represented using the planning domain
definition language or 
PDDL \cite{haslum2019introduction, mcdermott1998pddl} -- we will use
the same in this paper.
All the information to derive this solution (plan) is contained
in the input model which remains static during the planning task. 
{\em But what if the model itself needs to be changed?}

This may be because it is incorrect, 
or incomplete, or even
unsolvable. It may be because it needs to be changed
to support some new behaviors. 
It may also be because the model is being used to 
describe a world that itself needs to change through
the actions of an agent.
In practice, the deployment of systems that can plan
involves a whole gamut of challenges in authoring, maintaining,
and meta-reasoning about models of planning tasks. 
% \textcolor{red}{Refer to Figure \ref{fig:mod_space_problems} for a visual representation of classical planning and model reasoning problems.}

% \begin{itemize}
% % \item [-]
% % {\bf Code: } \url{https://github.com/HAPILab/modelmaker}
% \item [-]
% {\bf Supplementary files:} A longer version of the paper 
% % with detailed examples of prompts 
% is available at \url{https://arxiv.org/abs/2311.13720}.
% \end{itemize}

\begin{figure}[tbp!]
    \centering
    \includegraphics[width=\linewidth]{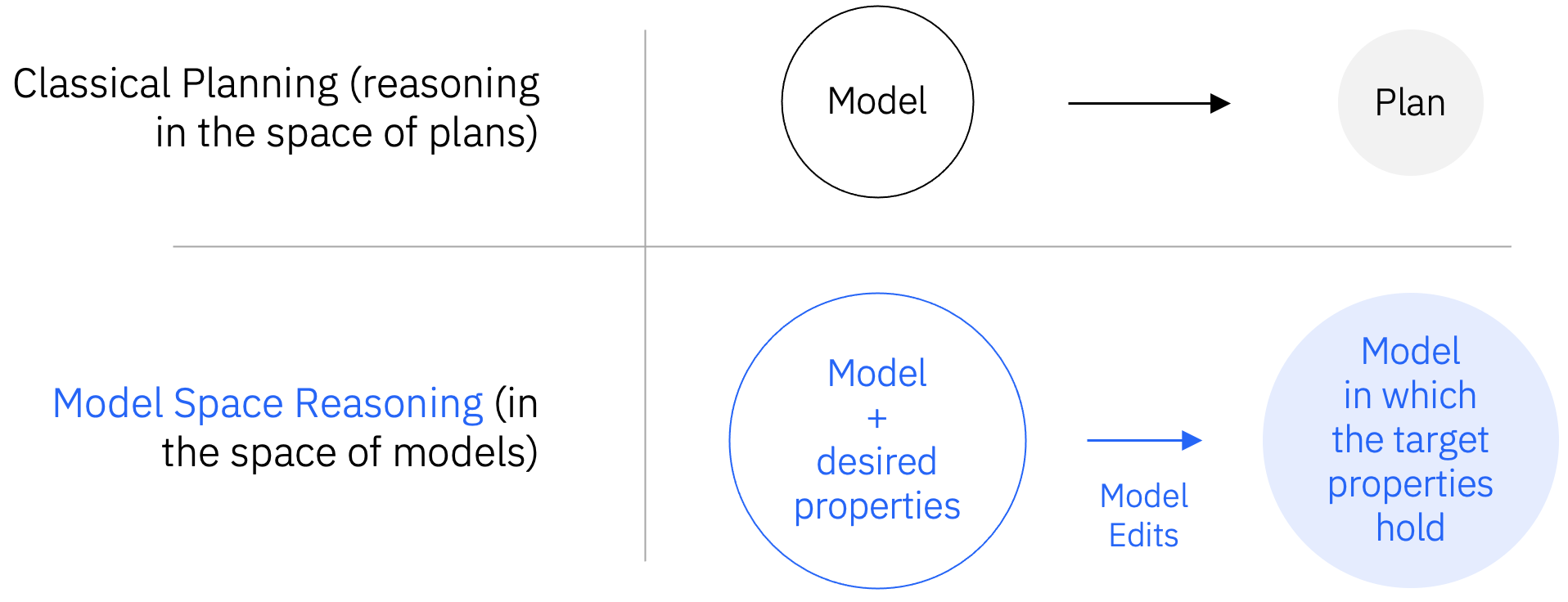}
    \caption{Classical planning versus model space problems.}
    \label{fig:mod_space_problems}
\end{figure}

\begin{figure*}
\centering
\begin{subfigure}[b]{0.275\textwidth}
\centering
\includegraphics[width=\textwidth]{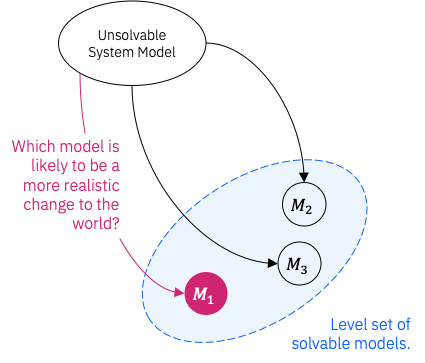}
\caption{Unsolvability.}
\label{fig:unsolvability}
\end{subfigure}
\hfill
\begin{subfigure}[b]{0.31\textwidth}
\centering
\includegraphics[width=\textwidth]{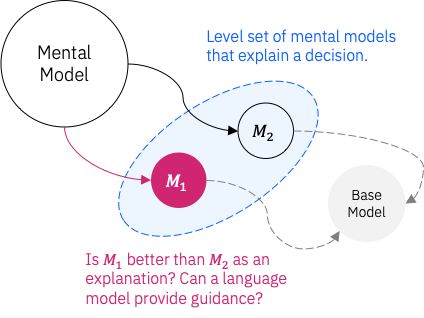}
\caption{Explanations}
\label{fig:explanations}
\end{subfigure}
\hfill
\begin{subfigure}[b]{0.3\textwidth}
\centering
\includegraphics[width=\textwidth]{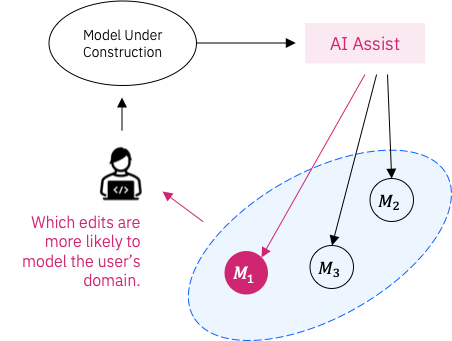}
\caption{Domain Authoring}
\label{fig:authoring}
\end{subfigure}
\caption{
A conceptual illustration of model space problems in AI planning.
Instead of the classical planning task of computing a plan given a model,
a model space task starts with a starting model $\mathcal{M}$ and a target criterion to satisfy,
and the solution is a new model $\mathcal{M}_1$ where that criterion is satisfied.
That criterion in Figure \ref{fig:unsolvability} is that the initially unsolvable model becomes
solvable (or an initially invalid plan in $\mathcal{M}$ becomes valid in the new model $\mathcal{M}_1$).
In Figure \ref{fig:explanations}, on the other hand, the starting model is the mental model of the user
that needs to be updated and the target is a new model that can explain a given plan (or refute 
a given foil). In domain authoring situations, such model updates happen with the domain writer
in the loop, and the starting model is the model under construction (Figure \ref{fig:authoring}). In all these cases, there are many non-unique 
model edits $\mathcal{M}_1 \Delta \mathcal{M}$ that can satisfy the required criterion. 
In this paper, we explore if LLMs can 
produce more likely edits in real-worldly domains.
}
\label{fig:overview}
\end{figure*}

\subsection*{Model Space Problems in AI Planning}
\label{subsec:flavors}

We begin by enumerating the different flavors of 
model space reasoning explored in the AI planning literature. 
All of them involve a starting model which has something wrong with it
and the solution is a new model where the problem has been resolved or the 
required criterion has been met (Figure \ref{fig:mod_space_problems}).
% For readers new to the subject, we provide a conceptual
% illustration of these topics in Figure \ref{fig:overview}.
% These will form the basis of the 
% rest of our study.

\subsubsection{Unsolvability}

Perhaps the most difficult of model space problems, especially with humans
in the loop, is that of unsolvability. This is because when 
a model is unsolvable, there is no artifact (such as an outputted plan)
to look at for debugging purposes.
While there have been a lot of efforts, including an ongoing
competition \cite{ipc-unsolvability}, to {\em detect} unsolvability
of planning tasks up-front to speed up calls to a planning module \cite{backstrom2013fast, moreira2017improving}, 
and attempts to compute or even learn
heuristics \cite{hoffmann2014distance, staahlberg2017tailoring, staahlberg2021learning} and 
produce certificates \cite{eriksson2017unsolvability, eriksson2018proof, eriksson2020certified}
for unsolvable tasks, 
to make this process as efficient as possible, these do not help to fix the 
issues with the model that make it unsolvable in the first place. 

One of the seminal works in this category \cite{gobelbecker2010coming} 
framed the problem as ``excuse generation'' where the authors
envisaged a reformulation of the input planning task 
where if only (i.e. an excuse) certain
things about the current state were changed then it
would become solvable.
In addition to initial state changes, this idea was later 
extended \cite{herzig2014revision} to cover other parts of the 
model and framed as a more general ``planning task revision'' problem.

While these works do not particularly consider a human in the loop,
authors in \cite{sreedharan-d3wa+, sreedharan2019can} 
have looked at the problem of explaining unsolvability of planning tasks to
users explicitly as a model evolution problem, using techniques like
domain abstractions (simplifications) to adjust to users with different
levels of expertise. Later efforts \cite{kasermachetli} 
have borrowed from these concepts and tried to operationalize them
% in the form of python packages
for developers.

\subsubsection{Executability}

While unsolvable models produce no plans, incorrect or incomplete models
produce wrong plans. Conversely, a desired plan may not be among
the best (or even valid) plans in a given model. 
This class of model evolution problems \cite{sreedharan-aaai-2020-expectation-aware, sreedharan-d3wa+, sreedharan2019can}
closely mimics the unsolvability problem but with an additional
input -- a plan -- that must be made valid in the target model.
Interestingly, since the given plan is not valid in the basis model,
the basis model together with the plan (i.e. a compiled model where both
are enforced) gets us back to the unsolvability situation above.
We will use this approach when we deal with this class of problems later in
this paper but, to be clear, we do treat it as a separate class of model 
space problems to study since the input involves a plan that a competent solver
% (in our case, an LLM)
must be able to reason about.

\subsubsection{Explanations} 

The above problems deal with one model in isolation.
However, when working with humans in the loop, AI systems are often
required to provide explanations of their behavior. 
Planning systems are no different
\cite{chakraborti-ijcai-2020-the-emerging-landscape, fox2017explainable, chakraborti-icaps-2019}.
The model evolution problem here involves reasoning explicitly with the 
model of the (system) explainer as the basis model and 
the mental model of the human (explainee) as the target model.
This task can be formulated as one of 
``model reconciliation'' \cite{chakraborti-ijcai-2017} -- 
an explanation is the model update that justifies a particular plan i.e.
if both models justify a plan then there is no need for explanations.
There is an overlap here with the previous tasks 
in terms of what kind of justifications
a user is looking for: it might be a justification for a plan that the system 
produced and is invalid in the user model, and we end up
in the unsolvability scenario again.
In the worst case, the system may have 
to refute all possible 
alternatives (called ``foils'' \cite{miller2019explanation})
and establish the optimality of a plan \cite{chakraborti-ijcai-2017}.
% This requirement makes the model space reasoning task for explanations
% distinct from the previous two tasks, 
% in addition to there being an explicit basis and target model for an explanation
% in contrast to open-ended model edits for the generic unsolvability and 
% executability tasks (Figure \ref{fig:overview}). 
% Hence, we study the effectiveness of an LLM on this task separately,
% as a distinct class of model space reasoning tasks.

Interestingly, one can 
remove \cite{chakraborti-icaps-2019-how} 
the basis model in the model reconciliation formulation and produce false 
explanations or ``lies''. While this makes for a computationally 
harder open-ended
search in the space of probable models, authors 
in \cite{chakraborti-icaps-2019-how} envisaged that algorithms which 
have looked at linguistic patterns for model 
evolution \cite{porteous2015automated, porteous2016planning} can assist in 
finding more probable models.
This, of course, raises several ethical questions \cite{chakraborti-aies}, 
especially now that LLMs can provide a stronger linguistic signal.
We do not study this task here for two reasons: 1) Technically, this is not
a separate class of a model reasoning problem since this ability 
is contained in the model reconciliation formulation; and 2) There seems to be
little reason for building systems that can lie more effectively.
% One could, however, potentially use the likelihood signal from LLMs to 
% detect lies effectively.

% \subsubsection{Lies} 

% Interestingly, and rather unfortunately, one can 
% remove \cite{chakraborti-icaps-2019-how} 
% the basis model in a model reconciliation formulation and produce false 
% explanations or ``lies''. While this makes for a computationally 
% harder open-ended
% search in the space of probable models, authors 
% in \cite{chakraborti-icaps-2019-how} envisaged that algorithms which 
% have looked at linguistic patterns for model 
% evolution \cite{porteous2015automated, porteous2016planning} can assist in 
% finding more probable models.
% This, of course, raises several ethical questions \cite{chakraborti-aies}, 
% especially now that LLMs can provide a much stronger linguistic signal.
% We do not study this task here for two reasons: 1) Technically, this is not
% a separate class of a model reasoning problem since this ability 
% is contained in the model reconciliation formulation; and 2) There seems to be
% little reason for building systems that can lie more effectively.
% One could, however, potentially use the probability signal from LLMs to 
% detect those lies effectively.

\subsubsection{Domain Authoring and Design}

While model evolution, in isolation, is useful for 
any autonomous system in a non-stationary domain, and 
explanations are a desired tool for any user-facing tool, 
a unique task in the context of planning systems 
we want to give a shout-out to is that of domain acquisition.
Planning requires models and a significant portion of those models
are acquired from domain experts. 
The knowledge acquisition literature in automated planning has 
studied this domain for decades \cite{vallati2020knowledge} 
and the difficulty of acquiring
domains remain a bottleneck in the adoption of planning technologies.

One subclass of domain authoring problems is {\em design} -- here,
the task is not to author a new domain but to evolve an existing one 
to optimize certain criteria like making the task of recognizing 
the goals of agents in the environment 
easier \cite{keren2014goal, mirsky2019goal, wayllace2016goal}
or making the behavior of agents easier to 
interpret \cite{kulkarni-icaps-2019, kulkarni-iros-2020}.
Here as well, search techniques reveal multiple possible design options 
that can be enforced on a domain to achieve the desired effect.
Issues of explanations, unsolvability, and executability
manifest themselves in domain authoring and design tasks,
with an additional component of interaction design
with the domain author in the loop.
Authors in \cite{sreedharan-d3wa+} demonstrate this in a
large-scale industrial domain on authoring models for goal-oriented
conversational agents \cite{muise-technical-2020}.
The role of an AI assist in authoring problems is especially 
critical in what we call ``real worldly domains''.

\subsection*{Real Worldly Domains and Likelihood of Models}
\label{subsec:real}

All the model space problems we talked about so far are usually
solved by some compilation to a combinatorial search process \cite{gobelbecker2010coming, chakraborti-ijcai-2017, sreedharan-aaai-2020-expectation-aware}
which terminates after a set of model edits satisfy the desired
properties in the modified model.
It is usually the case that this yields many non-unique solutions
-- e.g. there may be many explanations for the same plan, many ways to
change an unsolvable problem into a solvable one, or many ways to fix a model
in order to support an invalid plan.
From the perspective of a combinatorial search process, all these are 
logically equivalent and hence equally likely. 
In fact, in preliminary studies \cite{zahedi-hri-2019-towards}, 
it has already been demonstrated
how users perceive logically equivalent explanations generated through
a model reconciliation process, differently.

Large-scale statistical models such as LLMs, on the other hand,
carry a lot of domain knowledge on things we do in our everyday 
lives i.e. our worldly matters. 
For want of a better 
term\footnote{
While looking for a term to describe the domains describing our 
worldly matters, we overlooked two in particular.
In scientific literature, the term ``real-world domains''
is often used to establish something that is real but does 
come with an unnecessary connotation or snark of not being something of mere
academic interest aka a ``toy domain''. 
Furthermore, a so-called ``real world'' domain includes 
Mars rovers and unmanned vehicles, which are by no means part of
our worldly matters. On the other hand,
``common sense'' tasks are widely used to characterize 
things that come naturally to humans but our worldly matters can
involve much more complexity than common sense tasks -- e.g. a service
composition task -- and we do hope to find the knowledge of those activities
in the statistical signal from large-scale language models. 
We avoid both terms for these reasons but 
better suggestions are welcome.
}, 
we call these real worldly domains.
Broadly speaking, these include all manner of human enterprise -- and 
consequently (planning) models describing them wherever relevant (sequential
decision-making tasks) -- 
that are described on the public internet (and not the domain describing
the inner workings of a Mars rover
% or crew scheduling on a space station
per se).
Existing works leveraging LLMs for planning 
have already shown promising results in the classical
planning task in real worldly tasks in the home and kitchen 
\cite{ahn2022can, huang2022inner}, and in specialized but common 
tasks such as service 
composition \cite{langchain, semantic-kernel}.
Can LLMs do the same for model space reasoning for planning tasks?
Can LLMs give statistical insight into what model edits are more
likely when CS says they are equivalent? Can LLMs even bypass
the CS process, as it can in certain circumstances for the classical
planning task (Appendix Section B), 
{\em and do it all by itself??}
These are the questions we ponder in this work.

\subsubsection{Contributions}

% For several decades, the planning task has been approached 
% as one that can be solved
% using CS, and recently with the help of LLMs
% (more on this later in Section~\ref{sec:rw}).
% % As we mentioned in the beginning, 
% This is the first attempt at exploring
% the role of large-scale statistical models on model space search.
% To this end, 
% % we make two contributions:
% % 1) We enumerate with illustrative examples of how LLMs can assist different
% % flavors of model space reasoning tasks in AI planning.
% % 2) We 
% we analyze the role of an LLM
% in this process in terms of scalability and differentiating 
% properties -- particularly, for generating more likely model edits
% -- either in relation to CS as a direct replacement for 
% the model space reasoning task
% or in its role in an augmented approach with CS.

% This is the first attempt at exploring the role of LLMs on model space search.
This is the first attempt at an extensive and systematic exploration of the role of LLMs in model space search. To this end, we analyze the effectiveness of an LLM for generating more likely model edits
either in relation to CS as a direct replacement for the model space reasoning task or in its role in an augmented approach with CS. 

The answers to these questions have major implications beyond just an academic interest in
finding out the impact of LLMs on model space tasks in planning.
Unlike carefully crafted planning domains used as benchmarks, such as the ones used in the International Planning Competition (IPC) \cite{pd}, the deployment of planning models in real worldly domains has touchpoints with all the problems described above -- explainability of outputs and failure modes, investigation of unsolvability and executability in potentially faulty models, model authoring and maintenance over time, etc. -- often with the domain author in the loop \cite{sreedharan-aggregated, sreedharan-d3wa+}. These models are often not written by hand but generated on the fly at runtime from input data, either through code or using knowledge compilers like \cite{tarski:github:18}. An insight into the likelihood of models can empower the domain author to create and debug models with greater ease \cite{sreedharan-d3wa+, kasermachetli}, as well as allow automated model adaptation in fully autonomous systems in nonstationary environments \cite{bryce2016maintaining} or in constrained creative tasks like story-telling \cite{simon2022tattletale, porteous2016planning, porteous2021automated} that have previously relied on using limited linguistic cues like antonyms and synonyms \cite{porteous2015automated} for domain evolution.
% \textcolor{red}{Previous works have also looked at model correction using LLMs without explicit reference to model-space search. In particular \cite{gragera2023exploring} suggests that LLMs may not be effective in aiding model corrections. However, it’s important to note that this work was constrained by its focus on simpler and smaller-scale problems and did not take advantage of more advanced LLMs, like GPT-4.}

\begin{figure}
    \centering
    \includegraphics[width=\linewidth]{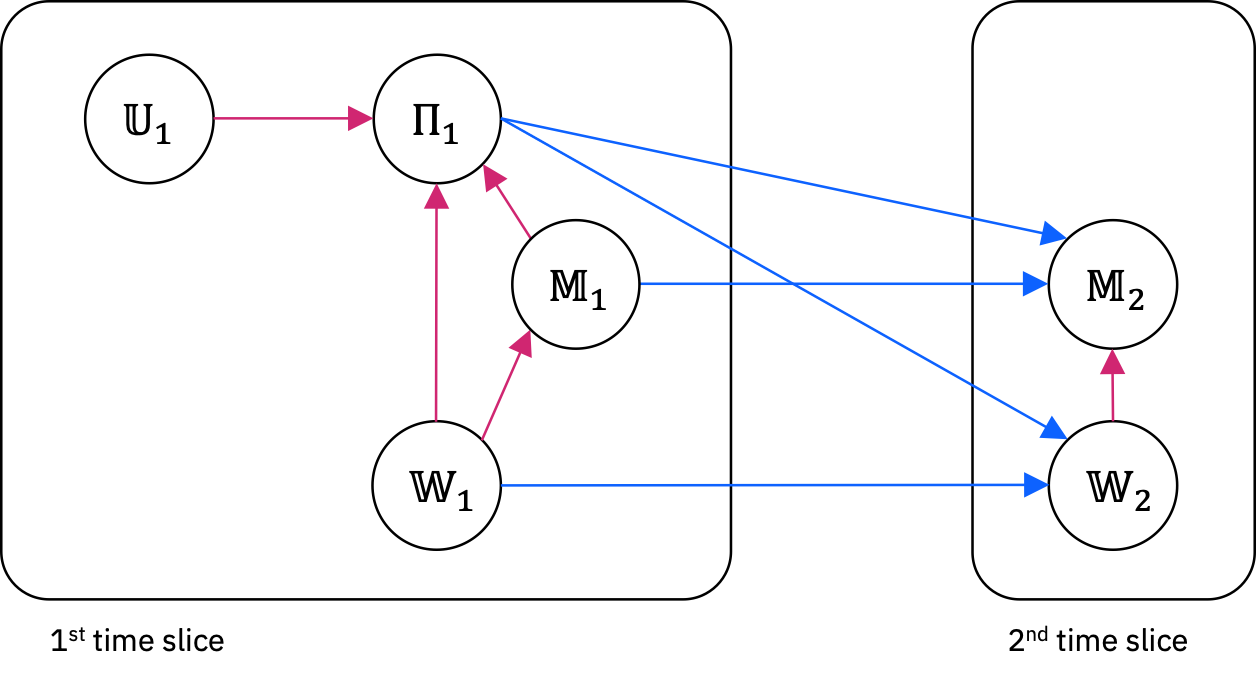}
    \caption{A DBN representing the random variables and their relations that are relevant to the problem at hand. The blue lines capture the diachronic, i.e., over time, relationships, and the maroon lines capture the synchronic ones.}
    \label{fig:prob}
\end{figure}

\section{Formal Interpretation of Model Likelihood}

In this section, we aim to provide a uniform probabilistic interpretation for the types of queries we employ in this problem. Figure \ref{fig:prob} presents a simplified dynamic Bayes network that encapsulates the scenario. This could be utilized to better comprehend and formalize the nature of the probabilities we intend to capture. Starting with the random variables, $\mathbb{M}_{1/2}$ and $\mathbb{W}_{1/2}$, these correspond to the model descriptions and the information about the true task/world at a given time step. The random variable $\Pi_i$ captures the policy that determines what action will be applied at a given step, which can alter the world and the model description. $\mathbb{U}_1$ determines the use case (this roughly maps to the type of model space search problem being solved).
The action combined with the use case, allows us to capture both scenarios where the focus is on updating the model description to better reflect the task (for example, domain authoring settings where the author may have misspecified something), and cases where the change also involves updating the underlying task and reflecting that change into the model description (for example, cases where the true task is unsolvable).
Please note that for explanation tasks, we expect  $\mathbb{M}_{1/2}$ to capture both the human knowledge about the task and the agent's model.

In the first time slice, we see that the actions that perform the update depend on the current model description, the task/world, and the use case. Naturally, this is a simplification of the true setting, but for the purpose of understanding the problem, this model serves as a useful abstraction. The most crucial term we are interested in measuring in this paper is the probability of an updated model description, given the prior model description and the use case:
\begin{equation}
  \label{eqn:basic}
  P(\mathbb{M}_{2}=\mathcal{M}_{2}\mid \mathbb{M}_{1}=\mathcal{M}_{1}, \mathbb{U}_{1}=\mathcal{U}).  
\end{equation}
We will examine cases where the information about $\mathbb{M}_{1}$ and $\mathbb{U}_{1}$ are included as part of the prompt, and we expect the LLM to approximate the above probability expression.

Note that this presupposes multiple capabilities of the LLM. For one, it assumes that the LLM can capture prior probabilities of possible world states. Next, it assumes that it can capture the likelihood of a specific action being performed for a given use case, state, and model description. Finally, it assumes that the LLM can discern how this action affects the next state and the model description. Furthermore, even if the LLM 
is capable of capturing this information separately, 
it may not correctly estimate the above probability expression. 
We hope to find a model such that:
\begin{align}
\begin{split}
\label{eqn:obj}
&\mathcal{M}  = \argmax_{\mathcal{M}^{'} \in \mathbb{M}} P(\mathbb{M}_{2}=\mathcal{M}^{'}\mid  \\[-3ex]
&~~~~~~~~~~~~~~~~~~~~~~~~~~~~~~~~~~~~\mathbb{M}_{1}=\mathcal{M}_{1}, \mathbb{U}_{1}=\mathcal{U}),
\end{split}
\end{align}
where $\mathbb{M}$ is the set of all possible model descriptions. 
% We will also have cases where we combine LLM's with a sound reasoner, in such cases we are effectively trying to find the following model
% \[\mathcal{M}  = \argmax_{\mathcal{M'} \in \hat{\mathbb{M}}} P(\mathbb{M}_{i+1}=\mathcal{M}_{i+1}\mid \mathbb{M}_{i}=\mathcal{M}_{i}, \mathbb{U}_{i}=\mathcal{U}),\]
% where $\hat{\mathbb{M}} \subseteq \mathbb{M}$, such that every model in $\hat{\mathbb{M}}$ meets the formal requirements to satisfy the use case $\mathcal{U}$. 
%It is also worth noting that in many of the queries that we consider, we are also including some information about the world state. \response{For example, in the unsolvability case, by providing a description and saying it's unsolvable, there is a strong possibility that this is a description that corresponds to the current world state.}

% \section{Proposed Modes of Leveraging LLMs for Model Space Exploration}
\section{LLMs ft. Model Space Exploration}
\label{sec:approaches}

In each of the model space search cases discussed before, we would ideally like to identify some model that satisfies Equation \ref{eqn:obj}.
However, to understand the current efforts in the model-space search, it might be useful to further decompose the metric into 
two components:

\begin{itemize}
\item {\bf Objective Metric}
This is the traditional metric that is being optimized by the various CS methods studied previously. In the cases we are focusing on, this is mostly a binary metric such as the solvability of a problem or the executability of the given plan. We will say a solution/model is {\em sound} if it satisfies the objective metric.

\item {\bf Likelihood of the Updated Model}
This is the specific aspect that is currently being overlooked by existing methods. This metric corresponds to the likelihood that the updated model generated through search corresponds to a desired target model. Equation \ref{eqn:basic} provides a formalization of this probability. The likelihood of different sound models would vary based on the use case and the context.

\end{itemize}

% In addition to these two metrics, there are other considerations such as the number or cost of model updates which are sometimes considered partial proxies for the latter criteria. We will ignore such metrics in this paper.

Our goal now is to find an updated model that meets the objective metric while maximizing its likelihood. As discussed, we will use pre-trained LLMs as the source for the information about the latter measure. One can envision four different configurations (see Figure \ref{fig:big-prob}) to achieve this goal:

\begin{figure*}[tbp!]
\centering
\includegraphics[width=0.925\linewidth]{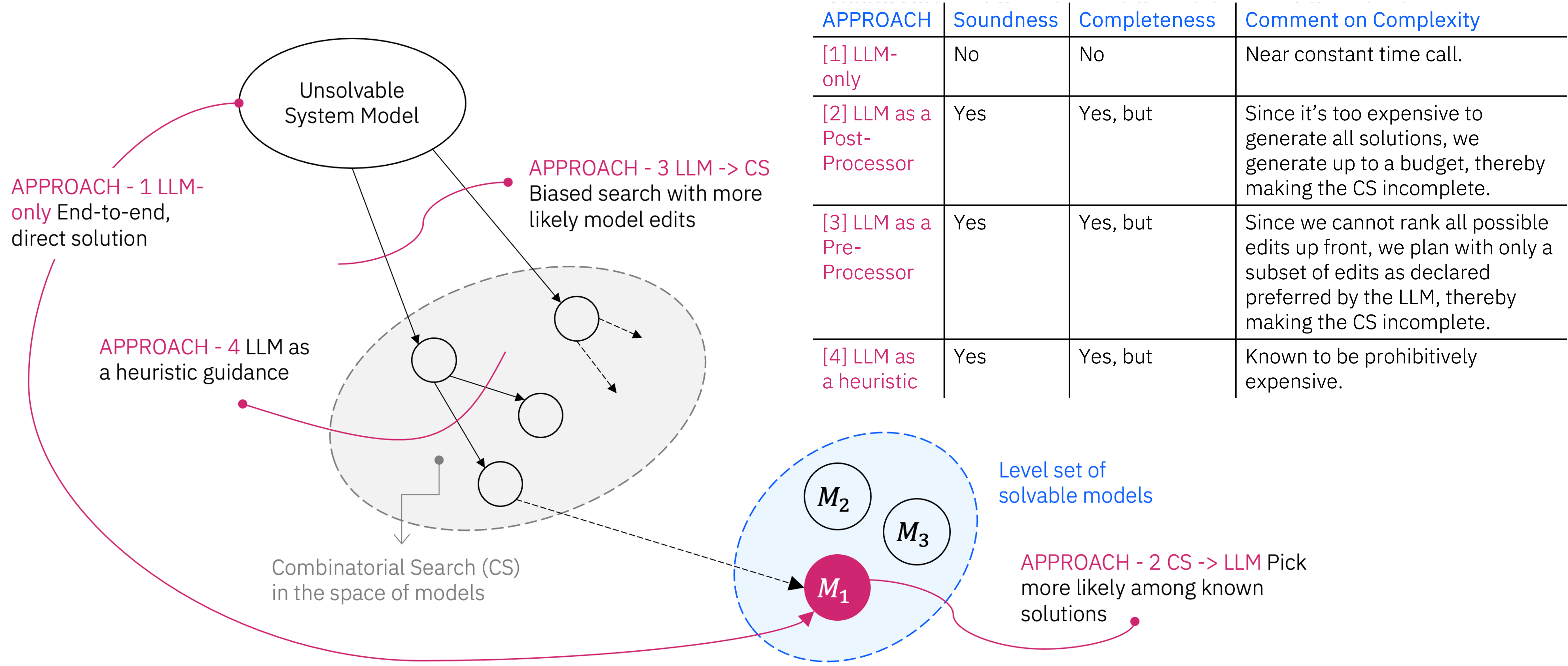}
\caption{Different points of contact with LLMs and the CS process.
While Approach-4 is known to be too expensive, 
we explore Approaches 1-3 in this paper in terms of the soundness and
likelihood of solutions.
}
\label{fig:big-prob}
\end{figure*}

\subsubsection{LLM-only Configuration}

In this mode, we provide the entire problem to LLM. The prompt is included with enough context that the system is aware of the criteria against which the likelihood of the models need to be measured. The LLM is asked to produce an updated model that is the most likely sound model. This corresponds to asking LLM to directly approximate Equation \ref{eqn:obj}.
We use the OpenAI API \cite{gpt-technical-2023} for this approach. 

\subsubsection{LLM as a Post Processor} 

In this mode, we use CS to generate a set of potential candidate solutions that are guaranteed to be sound. The LLM is then asked to select the model that is most likely. The prompt would again be designed to include the context necessary to determine what constitutes a target model. In this case, we are effectively trying to approximate the following problem:
\begin{align}
\begin{split}
&\mathcal{M}  = \argmax_{\mathcal{M}^{'} \in \hat{\mathbb{M}}} P(\mathbb{M}_{2}=\mathcal{M}^{'}\mid  \\[-3ex]
&~~~~~~~~~~~~~~~~~~~~~~~~~~~~~~~~~~~~\mathbb{M}_{1}=\mathcal{M}_{1}, \mathbb{U}_{1}=\mathcal{U}),
\end{split}
\end{align}
where $\hat{\mathbb{M}} \subseteq \mathbb{M}$, such that every model in $\hat{\mathbb{M}}$ meets the formal requirements to satisfy the use case $\mathcal{U}$.

Since enumerating all solutions is too expensive, 
we used an exhaustive search that caches solutions until a search budget of
5,000 (10,000) node expansions for unsolvability (inexecutability)
and a 2-hour limit was met per problem instance. 
This makes the solution incomplete.

\subsubsection{LLM as a Pre-Processor} 

In this mode, we ask the LLM to provide a ranked order of 
likely model edits without considering the objective metric. 
The ordering can then be used by CS to compute the most likely model 
that would satisfy or maximize the objective metric. 
% There are a few different options one could use here. One could ask the LLM to order purely based on the likelihood or based on both likelihood and the objective metric. 
This approach is still guaranteed to be sound, as the CS would 
only return a solution if the selected model updates result in a model 
that meets the objective metric. 
In this case, we are trying to approximate the following problem:
\begin{align}
\begin{split}
\label{eqn:post}
    \mathcal{M}  = \argmax_{\mathcal{M}^{'} \in \hat{\mathbb{M}}, \; \mathcal{M}^{'} \textrm{ is sound}} V(\mathcal{M}^{'}),
    % ,\\
    % \textrm{such that } \mathcal{M}^{'} \textrm{is sound.}
    \end{split}
\end{align}
where the utility/value function $V(\mathcal{M}^{'})$ is calculated from the 
LLMs approximation of the model likelihood.
Specifically, we will have $V(\mathcal{M}^{'}) \propto 
P(\mathbb{M}_{2}=\mathcal{M}^{'}\mid \mathbb{M}_{1}=\mathcal{M}_{1}, 
\mathbb{U}_{1}=\mathcal{U})$ if you are trying to order based on both 
objective metric and the likelihood of a model description, else you 
will have $V(\mathcal{M}^{'}) \propto P(\mathbb{M}_{2}=\mathcal{M}^{'}\mid 
\mathbb{M}_{1}=\mathcal{M}_{1})$. 

For the purposes of our implementation, we converted all the ordered edits proposed by the LLM into a set of actions that the CS can perform with different costs. In particular, we chose the cost of actions in such a way that, for an ordered sequence of $l$ edits, the total cost of including the first $i$ edits is always less than the cost of including the $i+1$th edit.
Since the LLM cannot rank all possible edits (capped at 20 for the experiments),
there is a possibility that the CS search will not be able to find a valid solution,
which makes this approach incomplete in practice as well.

\subsubsection{LLM for Search Guidance} 

This mode is particularly relevant if heuristic search is used. The search algorithm could leverage LLMs to obtain search guidance in the form of heuristic value. As with the previous mode, we can use LLM for getting information about both metrics and we can still guarantee correctness. The formal problem being approximated here again corresponds to the one listed in Equation \ref{eqn:post} and the value function considered will also have similar considerations.
This process requires calls to an LLM within the process of search and is known
to be \cite{ferber2020neural} computationally excessively prohibitive. 
Hence, we do not consider this configuration in our study.

In this paper, we focus primarily on evaluating two basic model space search problems, namely, addressing {\em unsolvability} and {\em plan executability}. The nature of the likelihood of the model could depend on the underlying use case in question.
One can broadly identify two classes of problems, namely 
{\em model misspecification} and {\em updating the environment}.
In the former case, the current model is misspecified and the model search is being employed to identify the true unknown underlying model.
In the latter case, the current model is an exact representation of the true environment, however the model and by extension the environment doesn't meet some desired properties. The goal here becomes to then identify the set of changes that can be made to the environment such that it meets the desired property. One could equivalently think of this being a case where there are actions missing from the model that correspond to these possible changes.
While both of these use cases have been considered in the literature, for simplicity the evaluation in the paper will primarily focus on the latter one. All prompts considered in the paper were written with the latter use case in mind.

% \begin{itemize}
%     \item[] {\bf Model misspecification} -- the current model is misspecified and the model search is being employed to identify the true unknown underlying model. 
%     \item[] {\bf Updating the environment} -- the current model is an exact representation of the true environment, however the model and by extension the environment doesn't meet some desired properties. The goal here becomes to then identify the set of changes that can be made to the environment such that it meets the desired property. One could equivalently think of this being a case where there are actions missing from the model that correspond to these possible changes.
% \end{itemize}

%%%New Unsolvability Table%%%%

\begin{table*}[!ht]
\small
\centering
\setlength\tabcolsep{2.75pt}
\begin{tabular}{l|rrrr|rrrr|rrrr}
\multicolumn{1}{l|}{\begin{tabular}[c]{@{}l@{}}
Unsolvability\end{tabular}} &
  \multicolumn{4}{c|}{LLM-Only} &
  \multicolumn{4}{c|}{LLM as Post Processor} &
  \multicolumn{4}{c}{LLM as Pre Processor} \\ 
 Domains &
  \multicolumn{2}{c|}{GPT-3.5-turbo} &
  \multicolumn{2}{c|}{GPT-4} &
  \multicolumn{2}{c|}{GPT-3.5-turbo} &
  \multicolumn{2}{c|}{GPT-4} &
  \multicolumn{2}{c|}{GPT-3.5-turbo} &
  \multicolumn{2}{c}{GPT-4} \\ \hline
 &
  \multicolumn{1}{l|}{Sound} &
  \multicolumn{1}{l|}{Preferred} &
  \multicolumn{1}{l}{Sound} &
  \multicolumn{1}{l|}{Preferred} &
  \multicolumn{1}{l|}{Solutions} &
  \multicolumn{1}{l|}{Preferred} &
  \multicolumn{1}{l|}{Solutions} &
  \multicolumn{1}{l|}{Preferred} &
  \multicolumn{1}{l|}{Ratio} &
  \multicolumn{1}{l|}{Preferred} &
  \multicolumn{1}{l|}{Ratio} &
  \multicolumn{1}{l}{Preferred} \\
Travel &
  \multicolumn{1}{r|}{97/245} &
  \multicolumn{1}{r|}{7/97} &
  164/245 &
  66/164 &
  \multicolumn{1}{r|}{245/245} &
  \multicolumn{1}{r|}{24/245} &
  \multicolumn{1}{r|}{245/245} &
  63/245 &
  \multicolumn{1}{r|}{129/245} &
  \multicolumn{1}{r|}{1/129} &
  \multicolumn{1}{r|}{160/245} &
  27/160 \\
Roomba &
  \multicolumn{1}{r|}{0/20} &
  \multicolumn{1}{r|}{0/0} &
  36/100 &
  7/36 &
  \multicolumn{1}{r|}{20/20} &
  \multicolumn{1}{r|}{2/20} &
  \multicolumn{1}{r|}{71/100} &
  9/71 &
  \multicolumn{1}{r|}{0/20} &
  \multicolumn{1}{r|}{0/0} &
  \multicolumn{1}{r|}{18/100} &
  4/18 \\
Logistics &
  \multicolumn{1}{r|}{61/69} &
  \multicolumn{1}{r|}{0/61} &
  65/69 &
  1/65 &
  \multicolumn{1}{r|}{69/69} &
  \multicolumn{1}{r|}{10/69} &
  \multicolumn{1}{r|}{69/69} &
  0/69 &
  \multicolumn{1}{r|}{56/69} &
  \multicolumn{1}{r|}{0/56} &
  \multicolumn{1}{r|}{65/69} &
  4/65 \\
Barman-S &
  \multicolumn{1}{r|}{43/61} &
  \multicolumn{1}{r|}{2/43} &
  57/61 &
  34/57 &
  \multicolumn{1}{r|}{34/61} &
  \multicolumn{1}{r|}{3/34} &
  \multicolumn{1}{r|}{34/61} &
  4/34 &
  \multicolumn{1}{r|}{28/61} &
  \multicolumn{1}{r|}{28/28} &
  \multicolumn{1}{r|}{17/61} &
  16/17 \\
Logistics-S &
  \multicolumn{1}{r|}{89/89} &
  \multicolumn{1}{r|}{0/75} &
  77/89 &
  28/77 &
  \multicolumn{1}{r|}{45/89} &
  \multicolumn{1}{r|}{3/45} &
  \multicolumn{1}{r|}{45/89} &
  5/45 &
  \multicolumn{1}{r|}{24/89} &
  \multicolumn{1}{r|}{0/24} &
  \multicolumn{1}{r|}{10/89} &
  5/10 \\
\textbf{Overall} &
  \multicolumn{1}{r|}{\textbf{276/484}} &
  \multicolumn{1}{r|}{\textbf{9/276}} &
  \textbf{399/564} &
  \textbf{136/399} &
  \multicolumn{1}{r|}{\textbf{194/484}} &
  \multicolumn{1}{r|}{\textbf{39/194}} &
  \multicolumn{1}{r|}{\textbf{198/564}} &
  \textbf{78/198} &
  \multicolumn{1}{r|}{\textbf{237/484}} &
  \multicolumn{1}{r|}{\textbf{29/237}} &
  \multicolumn{1}{r|}{\textbf{270/564}} &
  \textbf{56/270} \\ 
\end{tabular}
\caption{Results from the LLM-only, LLM as post-processor, and LLM as pre-processor settings for each unsolvability domain. 
% The "Overall" category represents the aggregated results.
}
\label{tab:tab_uns_merged}
% \end{minipage}
\end{table*}

%%%End of New Unsolvability Table%%%%

%%%% New Executability Table

\begin{table*}[!ht]
\small
\centering
\setlength\tabcolsep{3pt}
\begin{tabular}{l|rrrr|rrrr|rrrr}
\multicolumn{1}{l|}{\begin{tabular}[c]{@{}l@{}}
Executability\end{tabular}} &
  \multicolumn{4}{c|}{LLM-Only} &
  \multicolumn{4}{c|}{LLM as Post Processor} &
  \multicolumn{4}{c}{LLM as Pre Processor} \\
\multicolumn{1}{l|}{Domains} &
  \multicolumn{2}{c|}{GPT-3.5-turbo} &
  \multicolumn{2}{c|}{GPT-4} &
  \multicolumn{2}{c|}{GPT-3.5-turbo} &
  \multicolumn{2}{c|}{GPT-4} &
  \multicolumn{2}{c|}{GPT-3.5-turbo} &
  \multicolumn{2}{c}{GPT-4} \\ \hline
\multicolumn{1}{c|}{} &
  \multicolumn{1}{c|}{Sound} &
  \multicolumn{1}{c|}{Preferred} &
  \multicolumn{1}{c}{Sound} &
  \multicolumn{1}{c|}{Preferred} &
  \multicolumn{1}{c|}{Solutions} &
  \multicolumn{1}{c|}{Preferred} &
  \multicolumn{1}{c|}{Solutions} &
  \multicolumn{1}{c|}{Preferred} &
  \multicolumn{1}{c|}{Ratio} &
  \multicolumn{1}{c|}{Preferred} &
  \multicolumn{1}{c|}{Ratio} &
  \multicolumn{1}{c}{Preferred} \\
Travel &
  \multicolumn{1}{r|}{80/245} &
  \multicolumn{1}{r|}{33/80} &
  225/245 &
  130/225 &
  \multicolumn{1}{r|}{89/245} &
  \multicolumn{1}{r|}{38/89} &
  \multicolumn{1}{r|}{89/245} &
  57/89 &
  \multicolumn{1}{r|}{31/245} &
  \multicolumn{1}{r|}{31/31} &
  \multicolumn{1}{r|}{207/245} &
  207/207 \\
Roomba &
  \multicolumn{1}{r|}{0/20} &
  \multicolumn{1}{r|}{0/0} &
  57/99 &
  31/57 &
  \multicolumn{1}{r|}{12/20} &
  \multicolumn{1}{r|}{12/12} &
  \multicolumn{1}{r|}{16/99} &
  12/16 &
  \multicolumn{1}{r|}{0/20} &
  \multicolumn{1}{r|}{0/0} &
  \multicolumn{1}{r|}{67/99} &
  11/67 \\
Logistics &
  \multicolumn{1}{r|}{16/69} &
  \multicolumn{1}{r|}{0/16} &
  66/69 &
  11/66 &
  \multicolumn{1}{r|}{51/69} &
  \multicolumn{1}{r|}{5/51} &
  \multicolumn{1}{r|}{51/69} &
  22/51 &
  \multicolumn{1}{r|}{13/69} &
  \multicolumn{1}{r|}{2/13} &
  \multicolumn{1}{r|}{13/69} &
  20/57 \\
Barman-S &
  \multicolumn{1}{r|}{57/61} &
  \multicolumn{1}{r|}{14/57} &
  56/61 &
  15/56 &
  \multicolumn{1}{r|}{34/61} &
  \multicolumn{1}{r|}{8/34} &
  \multicolumn{1}{r|}{34/61} &
  13/34 &
  \multicolumn{1}{r|}{29/61} &
  \multicolumn{1}{r|}{29/29} &
  \multicolumn{1}{r|}{29/61} &
  26/26 \\
Logistics-S &
  \multicolumn{1}{r|}{21/89} &
  \multicolumn{1}{r|}{6/21} &
  89/89 &
  77/89 &
  \multicolumn{1}{r|}{68/89} &
  \multicolumn{1}{r|}{23/68} &
  \multicolumn{1}{r|}{68/89} &
  60/68 &
  \multicolumn{1}{r|}{0/89} &
  \multicolumn{1}{r|}{0/0} &
  \multicolumn{1}{r|}{0/89} &
  14/18 \\
\textbf{Overall} &
  \multicolumn{1}{r|}{\textbf{174/484}} &
  \multicolumn{1}{r|}{\textbf{53/174}} &
  \textbf{493/563} &
  \textbf{264/493} &
  \multicolumn{1}{r|}{\textbf{170/484}} &
  \multicolumn{1}{r|}{\textbf{32/170}} &
  \multicolumn{1}{r|}{\textbf{170/563}} &
  \textbf{110/170} &
  \multicolumn{1}{r|}{\textbf{73/484}} &
  \multicolumn{1}{r|}{\textbf{62/73}} &
  \multicolumn{1}{r|}{\textbf{375/563}} &
  \textbf{278/375} \\ 
\end{tabular}
    \caption{Results from the LLM-only, LLM as post-processor, and LLM as pre-processor settings for each executability domain. 
    % The "Overall" category represents the aggregated results.
    }
    \label{tab:tab_exec_merged}
\end{table*}

%%%% End of New Executability Table

\section{Empirical Results}

For evaluating the three approaches, we designed four novel domains
so that a certain set of changes would be
clearly recognized as more reasonable, i.e. more likely to 
be realized in the real world. 
We additionally assume that all changes that belong to this set 
(henceforth referred to as {\em ``reasonable changes''}),
will result in models with the same likelihood.
% All model updates were limited to initial state changes.
% The domains are the following:
%Hyperref is forbidden

\subsubsection{Travel Domain}
Here an agent travels from a given city to another, using
either a taxi or bus to travel between cities. 
We additionally encode which cities neighbor each other, 
and the initial problem only includes bus or taxi services between neighboring 
cities. Reasonable changes are limited to starting taxi or bus services 
between neighboring cities only.

\subsubsection{Roomba}

% In this domain, the agent needs to clean a specified room, which
% requires it to travel to the target room while traversing the intermediate 
% rooms through connecting doors. 
% The overall floor plan is laid out using neighboring predicates and the 
% expectation is there can only be doors between neighboring rooms. 
% Changes are reasonable if it only involves 
% whether a door is locked or unlocked (as opposed to connecting
% new rooms requiring removal of walls per se).

In this domain, the agent needs to clean a specified room, which requires it to travel to the target room while traversing the intermediate 
rooms through connecting paths. Along the paths, obstacles such as walls, chairs, or tables may be present. If a path is blocked, the agent can not move to an adjacent cell. Changes are reasonable if they involve removing chairs or tables that obstruct the path and adding `path clear' to the corresponding cells.

\subsubsection{Barman-simple}

This is a modified version of the IPC barman domain \cite{barman}. 
Here, the agent is expected to prepare a set of drinks, 
given a set of containers and ingredients. 
While only considering a subset of actions from the original domain, 
we introduce a new predicate that 
indicates whether a container is clean, which
is a precondition for using the container for a drink. 
We consider solutions to be reasonable if they only involve 
marking containers as clean (as opposed to
adding prepared drinks).

\subsubsection{Logistics-simple}

Finally, we consider a simplified version of the logistics problem
where a package is transported from one collection station to a target station. 
Each station contains a truck that can move the package to a neighboring station. 
We add a new precondition that ensures that only trucks that are marked as being
ready for transportation can be used to move packages. We limit 
reasonable changes to ones that mark trucks as being ready for transportation.

\subsection*{Experimental Setup}
\label{subsec:setup}

In each domain, we create a set of solvable problems of varying sizes. 
We then made it unsolvable by deleting a set of initial state predicates that correspond to reasonable changes. 
The number of such modifications ranges from 1 to 4. 
This means, by design, there exists a set of reasonable changes that can make the problem solvable. For the plan executability case, we chose one of the plans generated from the original solvable plan as the target plan to be made solvable.
All model updates were limited to initial state changes only.

\subsubsection{Phrasing of the prompts}

Our objective is to determine whether a model space solution is reasonable
in the sense of the likelihood of being realized in the real world. 
We captured this in the prompts by asking the LLM to generate or select the most {\em reasonable} set of model edits. 
We also tested with a more verbose prompt that explicitly mentions the ease of realizing the changes, more on this in Appendix Section C.

\subsubsection{Hypotheses} 

We focus on the following hypotheses,
for both the unsolvability and executability settings:

\begin{itemize}
\item[{\bf H1}] 
LLM can identify sound model updates.
\item[{\bf H2}] 
LLM can identify reasonable model updates.
\item[{\bf H3}] 
The ability to find sound model updates 
improves with the capability of the LLM.
\item[{\bf H4}] 
The ability to find reasonable model updates 
improves with the capability of the LLM.
\item[{\bf H5}] 
The ability to produce sound, and hence reasonable solutions
as a fraction of it, will be significantly outperformed 
by the two CS+LLM approaches.
\item[{\bf H6}] 
LLMs will provide a stronger signal, i.e. a higher fraction of 
sound and reasonable solutions, in public domains an LLM is likely to 
have seen already.
\item[{\bf H7}] 
The performance of an LLM will deteriorate with the complexity
of the model space reasoning task.
\end{itemize}

\begin{figure*}[!ht]
  \centering
  \begin{subfigure}[b]{0.49\textwidth}
    \includegraphics[width=\textwidth]{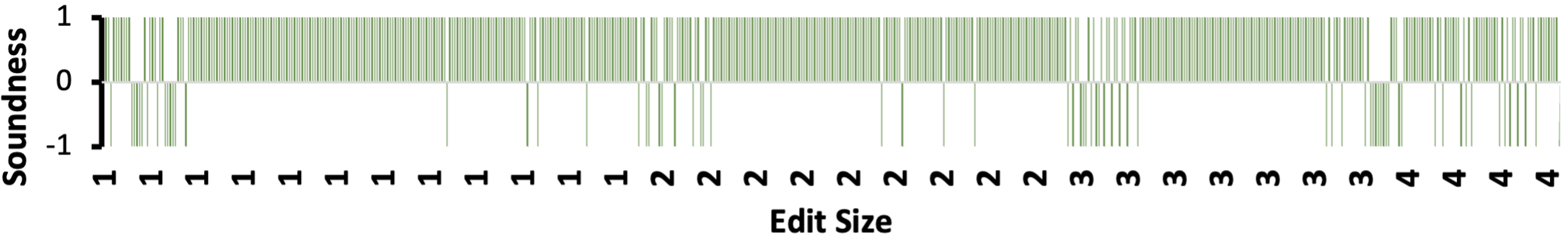}
    \caption{Unsolvability: Soundness vs. Edit size}
    \label{fig:image1}
  \end{subfigure}
  \hfill
  \begin{subfigure}[b]{0.49\textwidth}
    \includegraphics[width=\textwidth]{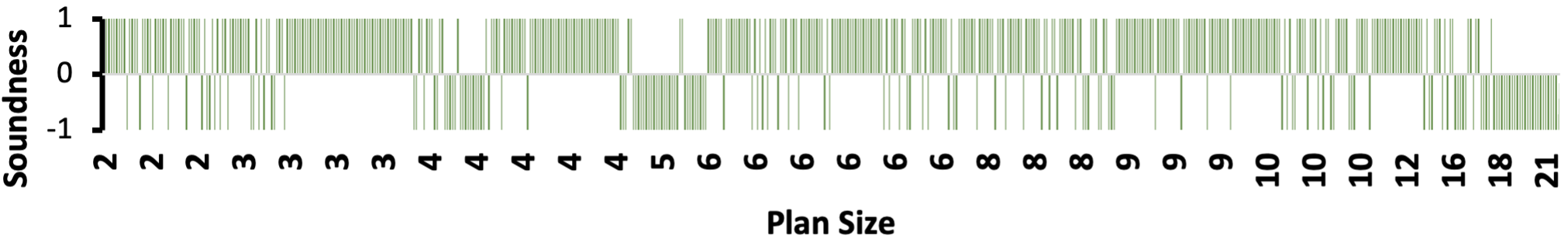}
    \caption{Unsolvability: Soundness vs. Plan size}
    \label{fig:image2}
  \end{subfigure}
  
  % \vspace{10pt}

  \begin{subfigure}[b]{0.49\textwidth}
    \includegraphics[width=\textwidth]{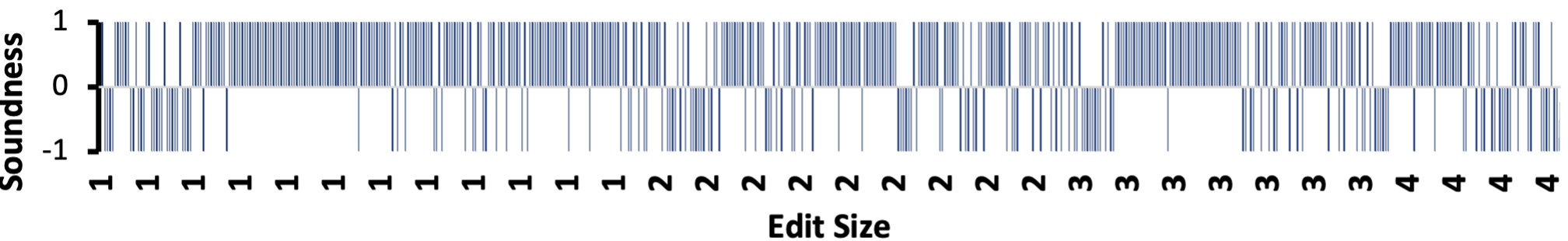}
    \caption{Executability: Soundness vs. Edit size}
    \label{fig:image3}
  \end{subfigure}
  \hfill
  \begin{subfigure}[b]{0.49\textwidth}
    \includegraphics[width=\textwidth]{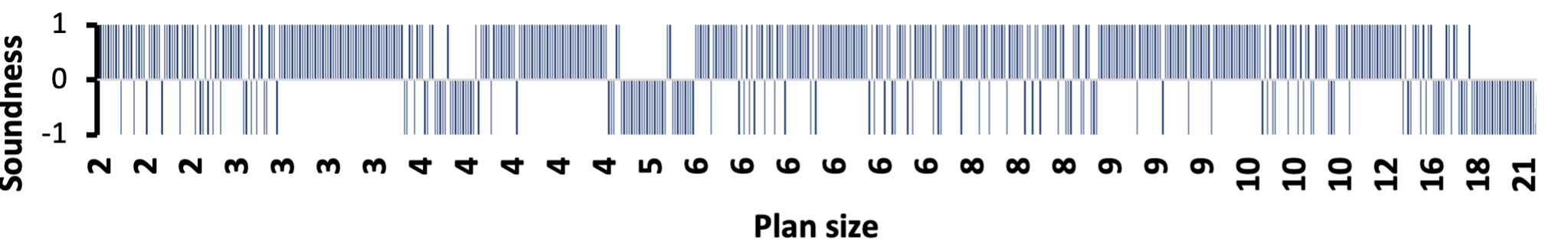}
    \caption{Executability: Soundness vs. Plan size}
    \label{fig:image4}
  \end{subfigure}
  \caption{Soundness of solutions from the LLM-only (GPT-4) approach against edit and plan sizes for unsolvability and executability settings in 564 problems across all 5 domains. Each bar represents one problem instance: a bar height of 1 indicates a sound solution, -1 otherwise. A higher concentration of negative bars will indicate deterioration in performance.
  }

  \label{fig:main}
\end{figure*}

\subsubsection{Measurements}

H1 and H2 are measured directly against the ground truth, as per the 
problem-generation process explained at the start of Section \ref{subsec:setup}.
For H3 and H4, we compare H1 and H2 from GPT-3.5-turbo to GPT-4.
For H5, we measure H1 and H2 relative to the two CS integrations
with the LLM as a pre-processor and LLM as a post-processor.
For H6, we compare H1-H4 in two ways: 1) the performance 
in two public domains Barman and Logistics, as compared to the 
two novel domains Travel and Roomba; and 2) the relative 
performance between Logistics and Logistics-simple, the latter being
a modified version of the former.
Finally, for H7, we measure how H1 and H2 fares with two measures
of complexity: 1) the number of model edits required to arrive at a solution;
and 2) the length of the plan underlying a model space reasoning task. 
For unsolvability, this is known when a planning task is made unsolvable
as per the problem generation process, while for executability,
the plan is part of the input to the reasoning task.

\subsection*{Results}

Tables \ref{tab:tab_uns_merged} and \ref{tab:tab_exec_merged} 
presents the outcomes for unsolvability and inexecutability setting
respectively. Since both display identical trends for H1-H7, 
we describe them together.
The only difference between the two settings is that the post-processing
approach had a larger budget for expanded nodes as mentioned in 
Section \ref{sec:approaches}, since it rarely hit the time budget. 
However, this did not make much difference.

In support of H1 and H2, the LLM-only approach
demonstrates surprising proficiency in suggesting 
sound and reasonable solutions across various domains. 
In support of H3-H4, the LLM-only approach sees the most pronounced improvement 
in identifying sound model alterations, accompanied by a higher rate
of reasonable solutions as well, as we upgrade to the latest LLM. 
The relative gain between sound and reasonable solutions
is slightly counter to expectations though,
since an LLM is supposed to be a stronger statistical signal on more likely
updates rather than a reasoner by itself.

This surprise carries onto the comparative results with CS+LLM approaches.
Contrary to H5, the LLM-only setting outperforms both CS+LLM approaches. 
Note that the CS+LLM approaches are guaranteed to be sound, so the deficit
in the ``solutions'' column is between a sound solution versus no solution at 
all (and not sound versus unsound solutions).
The only way we do not get a (sound) solution from the
LLM as a Post-Processor approach is if the CS stage does not terminate
within the time or memory budget (as mentioned in Section \ref{sec:approaches}).
Similarly, the two ways we do not get a solution for the LLM as a 
Pre-Processor approach is if the preferred set of reasonable edits from the LLM
are not sufficient for the CS to construct a solution, or as in the previous 
case, the search does not terminate.
While the CS+LLM approaches hit the computational curse, 
the LLM approach hits the curse of limited context size. 
Between GPT-3.5 and GPT-4, the prompt size has grown
from 4,096 to 8,192 tokens, but instances surpassing the token limit 
could not be processed. This makes a significant dent in the numbers
for the Roomba domain, especially for GPT-3.

The rate of sound solutions is much higher for public domains compared
to the custom ones, which is consistent with H6. 
However, this trend does not carry over to whether the solutions are 
reasonable or not. In fact, the derived logistics domain shows much 
higher rate of reasonable solutions than the public logistics domain
that shadows it. So results for H6 are inconclusive, and further 
underline the fickle nature of interfacing with LLMs.
Relatedly, the trends with respect to the complexity of the tasks, also
defy expectations. The rate of mistakes in constructing a sound solution
is spread uniformly across the spectrum of task complexity
(Figure \ref{fig:main}).

\section{Conclusion and Key Takeaways}

This is the first paper to consider the use of LLMs for 
model space reasoning tasks for automated planning. 
While the problem of model space search has been studied in various contexts, 
the question of how to evaluate the quality of different sound model updates
have mostly been left unanswered. 
Domain knowledge contained within LLM provides us with a powerful option
to evaluate the likelihood of different model updates. 
In contrast to early attempts \cite{gragera2023exploring} to use LLMs for model corrections, which were
constrained to limited settings and models that are no longer the state of the
art, we find LLMs to be surprisingly competent at this task.
In this paper, we exploited that power in 3 ways: first as a standalone
end-to-end approach and the others in conjunction with a sound solver.
The results reveal some intriguing trade-offs for the practitioner: 

\begin{itemize}
\item[-]
CS approaches are limited by the complexity of search.
Thus even while being theoretically sound and complete, they produce 
fewer solutions and hence fewer sound solutions in absolute numbers.
This means that augmenting the LLM-only approach with a validator \cite{howey2004val}
will produce as a whole a more effective sound and reasonable solution generator!

\item[-]
LLM approaches are limited by the size of the prompt and thus
does not scale to large domains even for computationally simpler
problem instances. 

\item[-]
The unpredictable nature of LLMs (e.g. H6 and H7) makes interfacing to LLMs
unreliable.
\end{itemize}

Despite these trade-offs, the promise of an LLM across H1-H5 is undeniable.
We are excited to explore further how this strong statistical signal
influences domain authoring tasks, as mentioned in Section \ref{subsec:real},
and reduces authoring overhead for planning tasks in the future.

% In this paper, we lay out a landscape of various model search problems that have been studied within the literature. We provide a general characterization of such model space search and see how one could employ LLM in the context of such problems. Our paper also presents some initial results on evaluating the effectiveness of using some state-of-the-art pre-trained LLM for some specific model space search problems. Our initial results are promising and also provide us with information about challenges related to using LLMs in this context. 
% Going forward, we would like to test our method on a larger variety of domains, use cases, and model search problem classes. 
%We would also like to look at testing other modes of using LLMs along with combinatorial search and see how they compare. We also hope to test our methods on other state-of-the-art LLMs and also see if the use of fine-tuning or more complex forms of prompting could improve the effectiveness of LLMs within this application. 

\section*{Acknowledgements}
Sarath Sreedharan’s research is supported in part by grant NSF 2303019.

% \section*{Appendix}

% A longer version of the paper 
% with detailed examples of prompts 
% is available at \url{https://arxiv.org/abs/2311.13720}.

\bibliography{bib}

\input{appendix}

\end{document}

%% file: appendix.tex
%%%%%%%%%%%%%%%%%%%%%%%%%%%%%%%%%%%%%%%%%%%%%%%%%%%%%%%%%%%%%%%%%%%%%%%%%%%%%%%
%%%%%%%%%%%%%%%%%%%%%%%%%%%%%%%%%%%%%%%%%%%%%%%%%%%%%%%%%%%%%%%%%%%%%%%%%%%%%%%
% APPENDIX
%%%%%%%%%%%%%%%%%%%%%%%%%%%%%%%%%%%%%%%%%%%%%%%%%%%%%%%%%%%%%%%%%%%%%%%%%%%%%%%
%%%%%%%%%%%%%%%%%%%%%%%%%%%%%%%%%%%%%%%%%%%%%%%%%%%%%%%%%%%%%%%%%%%%%%%%%%%%%%%
\newpage
\appendix
\onecolumn

\section{Limitations}

The proposed method has several limitations that need to be acknowledged. Firstly, its effectiveness is inherently limited by the capabilities of the LLMs it uses. As of the writing of the paper, LLMs have a number of known limitations that could prevent them from identifying the most likely models. Some of the issues include hallucination, lack of knowledge about specialized domains, the fact that it is an unsound reasoner, and so on (cf. \cite{valmeekam2023planning,bender2021dangers}). Secondly, it is currently hard to make the prediction generation more specific to a task or a user. This arises from various challenges, including practical limitations on fine-tuning the model to these specific settings or the inability to include all the relevant information in the prompt due to limitations in prompt size and context windows.

In Section \ref{subsec:flavors}, we noted the various flavors of model space problems in AI planning. We also noted how some of them overlap -- e.g. unsolvability, executability, and explanations in domain authoring tasks -- and how some of them are contained in others as  a strict subset -- e.g. explanations and lies.
In evaluating the proposed method, we only focused on two prominent use cases, namely unsolvability, and executability.
% In evaluating the proposed method we do not consider all the possible use cases, but instead chose to focus on two prominent ones, namely unsolvability, and executability.
Additionally, we only considered a specific type of model update namely adding new predicates into the initial state.
While theoretically, this model update can subsume any other model change, the ability of LLMs to identify likely model updates could differ based on the type of model updates considered.

Furthermore, the current study is limited to a set of domains where the reasonable or most likely changes were determined by the authors.
We limited testing to a few LLMs and only considered two of the four possible configurations.
%We also hope to te other modes of using LLMs along with combinatorial search and see how they compare. 
%We also hope to test our methods on other state-of-the-art LLMs and also see if the use of fine-tuning or more complex forms of prompting could  improve the effectiveness of LLMs within this application. 
%From that point of view, we ground our study in three specific
%problems: unsolvability, executability, and explanations, since
%the others are either a combination of these or are derived from these.
It is also worth noting that effective solutions for model space search may involve additional challenges that are not being evaluated here.
For example, domain authoring tasks also involve a human in the loop, which
introduces additional dimensions of study beyond just figuring 
out which model edits are more likely -- such as in figuring out how
to communicate those edits effectively to the domain author. 
Such considerations are out of the scope of this paper.
Similarly, the bastardized explainability problem that is able to generate lies, or conversely, a likely models approach that can actually catch those lies also have additional dimensions of interest, such as in mental modeling and computational ethics, which is also out of the scope of this work. 
%As the preliminary work in this direction, we will focus on two prominent ones, namely unsolvability, and executability.
We hope that this initial foray into this topic opens up future works in these directions.

%\section{Future Work}
In the future, we hope to address many of the limitations of the current evaluation. This includes expanding the number of use cases studied, considering various model updates, and comparing between all the possible configurations.
We will also look at the possibility of testing these methods in tasks, where we can correctly quantify the likelihood of these models. 
For unsolvability, this might involve focusing on scenarios where the cost of various actions possible in that setting is can be at least quantified accurately.
For use cases such as domain authoring, this might correspond to cases where the ground truth is known and as such one can correctly determine what the missing or incorrect model components could be.
We also hope to run user studies to evaluate the model updates generated by the method.

\section{Model Space Problems versus Other Meta-Reasoning Tasks in AI Planning}
\label{sec:model_vs_other}

% \section{Related Work}
% \label{sec:rw}

% In Section~\ref{subsec:flavors}, 
% we covered existing literature directly related to 
% model space reasoning for planning tasks.
% In the following, 
We include here some additional pointers to relevant
works that either explore the evolving role
of language models in planning or 
address other meta-reasoning tasks for planning.

\subsection*{Meta-Reasoning for Planning Tasks}

Reasoning about a planning model rather than using 
that model as immutable input to plan with, can be 
viewed as a form of meta-reasoning. 
Indeed, there is a long history of work on meta-reasoning
for planning tasks. However, these primarily involve a 
trade-off of the time taken to arrive at a solution 
versus the quality of the solution. 
Typically, in this setting, a planner can choose to stop
looking for better solutions, and potentially settle for 
a suboptimal solution, if it believes that there is 
(computationally) no point in carrying on.
Such approaches have been used
for policy optimization in Markov Decision
Processes \cite{lin2015metareasoning}, 
motion planning \cite{sung2021learning}, 
planning in temporal domains \cite{cserna2017planning}, 
heuristic search \cite{thayer2011learning}, and so on.
However, this thread of work does not aim to change
the model itself to better suit a given criterion, 
and that is our aim.

\subsection*{Model Space to State Space Compilations in Human-Aware Planning}

One meta-reasoning task that looks to change the 
model is ``human-aware planning'' -- this is explicitly formulated as a planning task of finding a plan \cite{chakraborti2018foundations}, and potentially some directive with it, given a basis model and
the mental model of the human(s) in the loop.
% ,
% again in contrast to the classical planning task of 
% finding a plan given a model.
In this paradigm, the directive accompanying the plan may be an update
to the mental model (i.e. an explanation of the plan).
In contrast to the traditional meta-reasoning approaches
that trade-off computation time with solution quality,
the reasoning task in human-aware planning trades off the solution
quality in the basis model with how it will be perceived in the 
mental model \cite{chakraborti-ijcai-2019-balancing}.

At this point, we want to make it clear that even though, conceptually, 
the model space reasoning problems described in this paper are looking
for solutions (new models) in the space of models, and classical planning tasks
are looking for solutions (plans) in the space of plans, 
these are not technically equivalent to plan-space and 
state-space search approaches 
used in planning \cite{ghallab2004automated}.
Indeed, if the reasoning task is compiled to be represented by a 
state-space representation, then both plans and models can be searched
for in the space of states. The approach 
in \cite{sreedharan-aaai-2020-expectation-aware} does exactly that 
for the explicability-explanations trade-off originally envisaged explicitly
in model-space search in \cite{chakraborti-ijcai-2019-balancing}. 
We do the same in our compilations for unsolvability and executability
for LLM as a pre-processor, while for LLM as a Post Processor,
we use the original model space search from \cite{chakraborti-ijcai-2017}.

\subsection*{Automated Planning ft. Neural Networks \& LLMs}
\label{subsec:planning-ft-llm}

Finally, there is a long history of work incorporating 
statistical models and machine learning, 
particularly deep neural networks, in planning tasks.
Historically, these works have only considered 
the classical planning task of computing a plan given a model.
To that end, researchers have looked at any and all aspects 
of the classical planning task of computing a plan given a model.

\subsubsection{Learning heuristics}

Heuristics play a key role in speeding up
CS by providing goal-directed guidance -- this
is typically achieved by solving a simplified version of the planning task 
at every search node and using information from that solution as guidance.
The better the approximation or simplification, 
the better the heuristic guidance.
As an alternative to the (human-made) approximation approach 
for devising heuristics, given experience from previous searches, 
one can train a model to learn a heuristic \cite{li2022optimal}.  
One of the early applications attempts to learn heuristics 
for automated planning using deep neural networks was 
in \cite{chen2011using}.
Many researchers \cite{shen2020learning, ferber2020neural, chrestien2021heuristic} 
have since tried to replicate this idea
with varying levels of success -- the computation overhead of a learned
heuristic during the search process remains an inhibiting factor. 

\subsubsection{Scaling up}

Automated planning, even in its simplest form, is computationally 
expensive \cite{ghallab2004automated}. 
Recent work \cite{groshev2018learning} 
have looked at training models on simpler problems
and using them to scale up to problems of higher complexity
where the learned approaches might not have all the nice guarantees 
of a traditional solver but, on the other hand, can at least solve
the problem with some level of quality instead of timing out. 
Relatedly, learning approaches can also be used to scale up planning
using model abstractions \cite{shah2022using}. 

\subsubsection{Transition functions}

There have been several 
attempts to learn the transition function of planning tasks 
in the form of PDDL directly from 
images \cite{asai2018classical, asai2019neural, asai2020learning} 
or text \cite{lindsay2017framer}.
There is an entire world of model learning for 
planning \cite{callanan2022macq} using both statistical as well as 
non-statistical learning techniques which we do not get into here.
Model learning as a task, although involves producing a model at the end,
is distinctly different from model space reasoning tasks in that the target
there is to produce a model that is maximum likelihood given an input 
dataset (versus evolving a given model to meet a set of desired properties).

\subsubsection{End-to-end}

Finally, with the increasing effectiveness of large-scale language models,
researchers are actively exploring whether a planning task can be 
done end-to-end using 
LLMs \cite{valmeekam2023planning, silver2022pddl, pallagani2022plansformer}.
Some recent approaches have even tried to produce PDDL representations
using large language models \cite{liu2023llm} 
-- not quite end-to-end but the final step from
PDDL to plan is lossless.
Perhaps one of the earliest approaches to using language models
in classical planning tasks is \cite{tian2015discovering}, 
where authors used word embeddings
to achieve a lower fidelity planning task they term ``plan completion''.
Follow-up works \cite{zhuo2020discovering} 
to this have also attempted to use other deep networks to this end.
The task of composing sequences, especially in the context 
of service composition using natural language as input, 
has also received much 
attention \cite{langchain, semantic-kernel}.
These are similar in fidelity to planning in real worldly
domains such as the one discussed previously 
in \cite{ahn2022can, huang2022inner}.
While still largely underwhelming in terms of the accuracy of the output
plans, these works do demonstrate rapid improvement. 
This is an intriguing development in planning research, especially 
as a way to bypass the computationally expensive combinatorial 
search process when possible.

\section{Prompt Variations}
\label{subsec:verbose}

As a way to test the effect the phrasing of our prompt had on the results, we also tried a variant of the prompt that was more explicit in what it expected to optimize for. Specifically, for generating a solvable problem variant we asked the system to:
{\em `Select the set of changes that would be the easiest to realize in the real world'}. Table \ref{tab:tab_explicit} shows the results of running this prompt for the LLM-only setting. The results can be compared directly to those presented in LLM-only columns of Table \ref{tab:tab_uns_merged}. The results are pretty similar, with the more verbose query being slightly worse off, so we do not explore  this direction in much more detail.

\begin{table}[bp!]
\centering
\small
\begin{tabular}{l|cc|cc}
\begin{tabular}[c]{@{}l@{}}Executibility\end{tabular} &
  \multicolumn{2}{c|}{GPT-3.5-turbo} &
  \multicolumn{2}{c}{GPT-4} \\ 
         Domains & \multicolumn{1}{c|}{Sound} & Preferred & \multicolumn{1}{c|}{Sound} & Preferred \\ \hline
Barman    & \multicolumn{1}{c|}{13/33} & 0/13      & \multicolumn{1}{c|}{33/33} & 27/33     \\
Logistics & \multicolumn{1}{c|}{17/25} & 0/17      & \multicolumn{1}{c|}{24/25} & 0/24      \\
\textbf{Overall} &
  \multicolumn{1}{c|}{\textbf{30/88}} &
  \textbf{0/30} &
  \multicolumn{1}{c|}{\textbf{57/58}} &
  \textbf{27/57} \\ 
\end{tabular}
\caption{The number of sound and reasonable model updates generated as a response to the more verbose query.}
\label{tab:tab_explicit}
\end{table}

\section{Sample Prompts}
\subsection{LLM only Setting for Unsolvability}
\begin{verbatim}
 Prompt Template:

 "given the following problem and domain files:" + domain_content + "," + 
 problem_content + "Come up with most reasonable set of additions that you 
 can make to the initial state that will make it solvable. I want you to only
 list the predicates to be added to the initial states without any 
 explanation or additional sentences in the beginning."

 Example values:
 
 domain_content:
    (define (domain domaingotocity)
    (:requirements :typing)
    (:types city - object)
    (:predicates 
        (at ?x - city)
        (has_taxi ?x ?y - city)
        (has_bus ?x ?y - city)
        (neighboring ?x ?y - city)
    )
    (:action use_taxi
        :parameters (?from ?to - city)
        :precondition (and 
          (at ?from) 
          (has_taxi ?from ?to) 
        )
        :effect (and
          (not (at ?from))
          (at ?to))
      )
    (:action use_bus
        :parameters (?from ?to - city)
        :precondition (and 
          (at ?from) 
          (has_bus ?from ?to)
        )
        :effect (and
          (not (at ?from))
          (at ?to))
      )
    )
    problem_content:
    (define (problem problemgotocity)
        (:domain domaingotocity)
        (:objects city_a - city 
        city_b - city city_c - city city_d - city 
        city_e - city city_f - city city_j - city 
        city_l - city city_o - city city_r - city 
        city_s - city city_t - city city_v - city 
        city_x - city)
        (:init 
        (at city_a) 
        (has_bus city_a city_b) 
        (has_bus city_a city_d) 
        (has_bus city_d city_j) 
        (has_bus city_l city_v) 
        (has_bus city_t city_e) 
        (has_taxi city_e city_o) 
        (has_taxi city_e city_x) 
        (has_taxi city_f city_s) 
        (has_taxi city_r city_l) 
        (has_taxi city_s city_c) 
        (neighboring city_a city_b) 
        (neighboring city_a city_d) 
        (neighboring city_b city_c) 
        (neighboring city_d city_j) 
        (neighboring city_e city_o) 
        (neighboring city_e city_x) 
        (neighboring city_f city_s) 
        (neighboring city_l city_v) 
        (neighboring city_r city_l) 
        (neighboring city_s city_c) 
        (neighboring city_t city_e) 
        (neighboring city_t city_o))
        (:goal (at city_c))
        )
    output:
    (:init (at city_a) 
    (has_bus city_a city_b) 
    (has_bus city_a city_d) 
    (has_bus city_d city_j) 
    (has_bus city_l city_v) 
    (has_bus city_t city_e) 
    (has_taxi city_e city_o) 
    (has_taxi city_e city_x) 
    (has_taxi city_f city_s) 
    (has_taxi city_r city_l) 
    (has_taxi city_s city_c) 
    (neighboring city_a city_b) 
    (neighboring city_a city_d) 
    (neighboring city_b city_c) 
    (neighboring city_d city_j) 
    (neighboring city_e city_o) 
    (neighboring city_e city_x) 
    (neighboring city_f city_s) 
    (neighboring city_l city_v) 
    (neighboring city_r city_l) 
    (neighboring city_s city_c) 
    (neighboring city_t city_e) 
    (neighboring city_t city_o) 
    (at city_o) 
    (at city_x))
\end{verbatim}
\clearpage

\subsection{LLM as Post Processor for Unsolvability}

\begin{verbatim}
 Prompt Template:

 "Given the following problem, domain files, and options list:
 - Problem: {uns_problem_string}
 - Domain: {domain_string} 
 - Options: {option_list} 
 Pick the most reasonable option from the list that you can apply to the 
 initial state to make the problem solvable. Only provide the number of 
 the option selected and no other information (exclude even the term 
 option)."
 
 Example values:

    uns_problem_string:
    (define (problem problemgotocity)
        (:domain domaingotocity)
        (:objects city_a - city 
        city_b - city city_c - city
        city_d - city city_e - city 
        city_f - city city_j - city 
        city_l - city city_o - city 
        city_r - city city_s - city 
        city_t - city city_v - city 
        city_x - city)
        (:init (at city_a) 
        (has_bus city_a city_b) 
        (has_bus city_a city_d) 
        (has_bus city_d city_j) 
        (has_bus city_l city_v) 
        (has_bus city_t city_e) 
        (has_taxi city_e city_o) 
        (has_taxi city_e city_x) 
        (has_taxi city_f city_s) 
        (has_taxi city_r city_l)
        (has_taxi city_s city_c) 
        (neighboring city_a city_b)
        (neighboring city_a city_d)
        (neighboring city_b city_c)
        (neighboring city_d city_j)
        (neighboring city_e city_o)
        (neighboring city_e city_x)
        (neighboring city_f city_s)
        (neighboring city_l city_v)
        (neighboring city_r city_l)
        (neighboring city_s city_c)
        (neighboring city_t city_e)
        (neighboring city_t city_o))
        (:goal (at city_c))
    )
    domain_string:
    (define (domain domaingotocity)
        (:requirements :typing)
        (:types city)
        (:predicates (at ?x - city)
        (has_bus ?x - city ?y - city)
        (has_taxi ?x - city ?y - city)
        (neighboring ?x - city 
        ?y - city))
        (:action use_bus
            :parameters 
            (?from - city ?to - city )
            :precondition 
            (and (at ?from) 
            (has_bus ?from ?to))
            :effect 
            (and (not (at ?from))
            (at ?to))
        )
         (:action use_taxi
            :parameters 
            (?from - city ?to - city )
            :precondition (and (at ?from) 
            (has_taxi ?from ?to))
            :effect (and (not (at ?from))
            (at ?to))
        )
    )
    
    options:
    
    ["Option 1: {'has_taxi city_a city_c'}",
    "Option 2: {'has_taxi city_b city_c'}",
    "Option 3: {'has_bus city_d city_c'}", 
    "Option 4: {'has_taxi city_a city_s'}",
    "Option 5: {'has_bus city_a city_s'}",
    "Option 6: {'has_bus city_b city_c'}",
    "Option 7: {'has_taxi city_b city_s'}",
    "Option 8: {'has_taxi city_d city_s'}",
    "Option 9: {'has_bus city_b city_f'}",
    "Option 10: {'at city_s'}",
    "Option 11: {'has_bus city_a city_c'}",
    "Option 12: {'has_bus city_a city_f'}",
    "Option 13: {'has_taxi city_j city_s'}",
    "Option 14: {'has_bus city_j city_f'}", 
    "Option 15: {'has_bus city_d city_f'}",
    "Option 16: {'has_taxi city_d city_c'}",
    "Option 17: {'has_taxi city_j city_f'}",
    "Option 18: {'at city_f'}",
    "Option 19: {'has_bus city_d city_s'}",
    "Option 20: {'has_bus city_j city_s'}"]
        
    output:
    4

\end{verbatim}
\clearpage

\subsection{LLM as Pre Processor Setting for Unsolvability}
\begin{verbatim}

Prompt Template:

prompt = f"Given the following problem and domain file: 
Problem:{uns_problem_string} Domain:\n {domain_string}
Come up with a list of twenty predicates that are currently
missing from the initial state. Order the predicates in such
a way that the predicates in the top correspond to changes that
are most reasonable to make (the predicate will added to the
existing initial state). Only list the initial state predicate, 
one predicate in a line, and provide no other information. 
Do not include any number in the list and do not include any
text before the list."

Example values:

    uns_problem_string:
    (define (problem problem_barman)
    (:domain barman)
    (:objects cocktail_a - cocktail 
    ingredient_a - ingredient 
    ingredient_b - ingredient 
    shaker_a - shaker shot_a - shot)
    (:init (cocktail-part1 c
    ocktail_a ingredient_a) 
    (cocktail-part2 cocktail_a ingredient_b)
    (contains shaker_a ingredient_a)
    (contains shaker_a ingredient_b) 
    (empty shot_a) (unshaked shaker_a))
    (:goal (contains shot_a cocktail_a))
    )

domain_string:
 (define (domain barman)
	(:requirements :strips :typing)
	(:types
		beverage container - object
		ingredient cocktail - beverage
		shot shaker - container
	)
	(:predicates
		(empty ?c - container)
		(contains ?c - container ?b - beverage)
		(clean ?c - container)
		(unshaked ?s - shaker)
		(shaked ?s - shaker)
		(cocktail-part1 ?a - cocktail ?b - ingredient)
		(cocktail-part2 ?a - cocktail ?b - ingredient)
	)

	(:action shake
		:parameters (?b - cocktail ?d1 ?d2 - ingredient ?s - shaker)
		:precondition (and
			(contains ?s ?d1)
			(contains ?s ?d2)
			(unshaked ?s))
		:effect (and
			(not (unshaked ?s))
			(not (contains ?s ?d1))
			(not (contains ?s ?d2))
			(shaked ?s)
			(cocktail-part1 ?b ?d1)
			(cocktail-part2 ?b ?d1)
			(contains ?s ?b))
	)

	(:action pour-shaker-to-shot
		:parameters (?b - cocktail ?d - shot ?s - shaker ?d1 ?d2 - ingredient)
		:precondition (and
			(shaked ?s)
			(empty ?d)
			(clean ?d)
			(contains ?s ?b)
			(cocktail-part1 ?b ?d1)
			(cocktail-part2 ?b ?d2)
		)
		:effect (and
			(not (clean ?d))
			(not (empty ?d))
			(contains ?d ?b)
		))
)

output:
[(empty shaker_a),
(empty cocktail_a),
(clean shaker_a),
(clean shot_a),
(contains cocktail_a ingredient_b),
(contains shaker_a shot_a),
(contains shaker_a cocktail_a),
(unshaked shaker_a),
(cocktail-part1 cocktail_a ingredient_b),
(cocktail-part2 cocktail_a ingredient_a),
(contains shot_a ingredient_a),
(contains shot_a ingredient_b),
(cocktail-part1 cocktail_a ingredient_b),
(cocktail-part2 cocktail_a ingredient_a),
(contains shaker_a cocktail_a),
(contains shaker_a ingredient_a),
(contains shaker_a ingredient_b),
(contains shot_a cocktail_a),
(clean cocktail_a),
(unshaked shot_a)]

\end{verbatim}
\clearpage

\subsection{LLM only Setting for Executability}
\begin{verbatim}

 Prompt Template:

"given the following problem and domain and plan files:" + domain_content + ","
+ problem_content + ","+ plan_content + "," + "Come up with most reasonable set 
of additions and deletes that you can make to the initial state to make the plan 
executable.I want you to list two sets of predicates 1) predicates to be added
to the initial states 2) predicates to be removed from the initial states.
Give me the predicates without any explanation or additional sentences 
in the beginning."

 Example values:
 
domain_content:
(define (domain cleaning)
(:requirements :typing)
(:types room door -object)
(:predicates
    (connects ?x - room ?y - room ?z - door)
    (is_open ?x - door)
    (at ?x - room)
    (is_dirty ?x - room)
    (is_unlocked ?x - door)
    (neighboring ?x ?y - room)
    (is_clean ?x - room)
)
(:action open_door
    :parameters (
    ?x - door
    )
    :precondition (and
      (is_unlocked ?x)
    )
    :effect (and
      (is_open ?x)
    )
)
(:action go
    :parameters (
    ?from - room
    ?to - room
    ?x - door
    )
    :precondition (and
      (at ?from)
      (connects ?from ?to ?x)
      (is_open ?x)
      (neighboring ?from ?to)
    )
    :effect (and
      (at ?to)
      (not(at ?from))
    )
)
(:action clean
    :parameters (
    ?x - room
    )
    :precondition (and
      (at ?x)
      (is_dirty ?x)
    )
    :effect (and
      (is_clean ?x)
    )
)
)

problem_content:
(define (problem problem_cleaning_robot)
    (:domain cleaning)
    (:objects door_a - door door_b - door door_c - door door_d - door 
    room_a - room room_b - room room_c - room room_d - room room_e - room 
    room_f - room room_g - room room_h - room)
    (:init (at room_a) (connects room_a room_a door_b) 
    (connects room_a room_b door_a) (connects room_a room_c door_d)
    (connects room_a room_d door_c) (is_dirty room_b) 
    (neighboring room_a room_b) (neighboring room_a room_h) 
    (neighboring room_c room_d) (neighboring room_d room_g)
    (neighboring room_f room_a) (neighboring room_f room_d) 
    (neighboring room_h room_e))
    (:goal (is_clean room_b))
)
plan_content:
(open_door door_a )
(go room_a room_b door_a )
(clean room_b )

output:
1) Predicates to be added to the initial states:
(path_is_clear cell_0_0 cell_1_0)
(path_is_clear cell_1_0 cell_1_1)

2) Predicates to be removed from the initial states:
(chair_blocking_path_between cell_0_0 cell_1_0)
(chair_blocking_path_between cell_1_0 cell_1_1)

\end{verbatim}
\clearpage
\subsection{LLM as Post Processor for Executability}
\begin{verbatim}

 Prompt Template:

 "Given the following problem, domain files, and options list:
    - Problem: {uns_problem_string}
    - Domain: {domain_string}
    - Options: {option_list}

Pick the most reasonable option from the list that you can apply to the initial 
state to make the following plan executable. 
- Plan: {original_plan} 
Only provide the number of the option selected and no other information 
(exclude even the term option)."

 Example values:

    uns_problem_string:
    (define (problem problemgotocity)
        (:domain domaingotocity)
        (:objects city_a - city city_b - city city_c - city city_d - city 
        city_e - city city_f - city city_g - city city_h - city city_i - city 
        city_j - city)

        (:init (at city_a) (has_bus city_a city_b) (has_bus city_a city_f) 
        (has_bus city_a city_i) (has_bus city_a city_j) 
        (has_bus city_h city_g) (has_bus city_j city_f) (has_taxi city_c city_d)
        (has_taxi city_c city_e) (has_taxi city_d city_e) 
        (has_taxi city_g city_f) (has_taxi city_h city_a) 
        (has_taxi city_j city_a) (neighboring city_a city_b)
        (neighboring city_a city_f) (neighboring city_a city_i) 
        (neighboring city_a city_j) (neighboring city_b city_c) 
        (neighboring city_c city_d) (neighboring city_c city_e) 
        (neighboring city_d city_e) (neighboring city_f city_c) 
        (neighboring city_g city_f) (neighboring city_h city_a) 
        (neighboring city_h city_g) (neighboring city_j city_a) 
        (neighboring city_j city_f))
        (:goal (at city_e))
    )

    domain_string:
        (define (domain domaingotocity)
    (:requirements :typing)
    (:types city - object)
    (:predicates 
        (at ?x - city)
        (has_taxi ?x ?y - city)
        (has_bus ?x ?y - city)
        (neighboring ?x ?y - city)
    )
      
      (:action use_taxi
        :parameters (?from ?to - city)
        :precondition (and 
          (at ?from) 
          (has_taxi ?from ?to) 
        )
        :effect (and
          (not (at ?from))
          (at ?to)
        )
      )
    
      (:action use_bus
        :parameters (?from ?to - city)
        :precondition (and 
          (at ?from) 
          (has_bus ?from ?to)
        )
        :effect (and
          (not (at ?from))
          (at ?to)
        )
      )
    )

    option_list:
    ['Option 1: (has_bus city_b city_c)', 
    'Option 2: (has_bus city_b city_c) (has_bus city_c city_d)',
    'Option 3: (has_bus city_b city_c) (has_bus city_b city_i)',
    'Option 4: (has_bus city_b city_c) (has_bus city_g city_d)',
    'Option 5: (has_bus city_g city_i) (has_bus city_b city_c)',
    'Option 6: (has_bus city_b city_c) (has_bus city_h city_c)',
    'Option 7: (has_taxi city_e city_d) (has_bus city_b city_c)',
    'Option 8: (has_bus city_b city_c) (has_taxi city_f city_j)',
    'Option 9: (has_taxi city_h city_i) (has_bus city_b city_c)',
    'Option 10: (has_bus city_b city_c) (neighboring city_b city_i)',
    'Option 11: (has_bus city_b city_c) (neighboring city_c city_b)',
    'Option 12: (has_bus city_b city_c) (has_bus city_c city_f)',
    'Option 13: (has_bus city_d city_d) (has_bus city_b city_c)',
    'Option 14: (has_bus city_b city_c) (has_bus city_e city_g)',
    'Option 15: (has_bus city_b city_c) (has_bus city_f city_a)',
    'Option 16: (has_bus city_b city_c) (has_bus city_f city_h)',
    'Option 17: (has_bus city_b city_c) (has_bus city_g city_e)',
    'Option 18: (has_bus city_b city_c) (has_bus city_j city_c)',
    'Option 19: (has_bus city_j city_i) (has_bus city_b city_c)',
    'Option 20: (has_bus city_b city_c) (has_taxi city_a city_b)']

    original_plan:
    (use_bus city_a city_b)
    (use_bus city_b city_c)
    (use_taxi city_c city_e)
    ; cost = 3 (unit cost)

    output:
    1
\end{verbatim}
\clearpage

\subsection{LLM as Pre Processor Setting for Executability}
\begin{verbatim}

 Prompt Template:

 prompt = f"Given the following problem, domain, and plan file:  
 Problem: {uns_problem_string} 
 Domain: {domain_string}
 Plan: {solv_init_plan_string} 
 Come up with a list of twenty predicates that are currently missing from the initial state
 to make the plan executable. Order the predicates in such a way that the predicates in
 the top correspond to changes that are most reasonable to make (the predicate will added 
 to the existing initial state). Only list the initial state predicate, one predicate in a
 line, and provide no other information. Do not include any number in the list and do
 not include any text before the list."

 Example values:

    uns_problem_string:
    (define (problem problem_barman)
    (:domain barman)
    (:objects cocktail_a - cocktail 
    ingredient_a - ingredient 
    ingredient_b - ingredient 
    shaker_a - shaker shot_a - shot)
    (:init (cocktail-part1 c
    ocktail_a ingredient_a) 
    (cocktail-part2 cocktail_a ingredient_b)
    (contains shaker_a ingredient_a)
    (contains shaker_a ingredient_b) 
    (empty shot_a) (unshaked shaker_a))
    (:goal (contains shot_a cocktail_a))
    )

    domain_string:
    (define (domain barman)
	(:requirements :strips :typing)
	(:types
		beverage container - object
		ingredient cocktail - beverage
		shot shaker - container
	)
	(:predicates
		(empty ?c - container)
		(contains ?c - container ?b - beverage)
		(clean ?c - container)
		(unshaked ?s - shaker)
		(shaked ?s - shaker)
		(cocktail-part1 ?a - cocktail ?b - ingredient)
		(cocktail-part2 ?a - cocktail ?b - ingredient)
	)

	(:action shake
		:parameters (?b - cocktail ?d1 ?d2 - ingredient ?s - shaker)
		:precondition (and
			(contains ?s ?d1)
			(contains ?s ?d2)
			(unshaked ?s))
		:effect (and
			(not (unshaked ?s))
			(not (contains ?s ?d1))
			(not (contains ?s ?d2))
			(shaked ?s)
			(cocktail-part1 ?b ?d1)
			(cocktail-part2 ?b ?d1)
			(contains ?s ?b))
	)

	(:action pour-shaker-to-shot
		:parameters (?b - cocktail ?d - shot ?s - shaker ?d1 ?d2 - ingredient)
		:precondition (and
			(shaked ?s)
			(empty ?d)
			(clean ?d)
			(contains ?s ?b)
			(cocktail-part1 ?b ?d1)
			(cocktail-part2 ?b ?d2)
		)
		:effect (and
			(not (clean ?d))
			(not (empty ?d))
			(contains ?d ?b)
		))
        )

    solv_init_plan_string:
    (shake cocktail_a ingredient_a ingredient_b shaker_a)
    (pour-shaker-to-shot cocktail_a shot_a shaker_a ingredient_a ingredient_a)
    ; cost = 2 (unit cost) 

output:

[(empty shot_e),
(empty shot_f),
(clean shot_c),
(clean shot_d),
(contains shaker_a ingredient_c),
(contains shaker_a ingredient_d),
(contains shaker_b ingredient_a),
(contains shaker_b ingredient_d),
(contains shaker_c ingredient_a),
(contains shaker_c ingredient_b),
(contains shaker_d ingredient_a),
(contains shaker_d ingredient_b),
(contains shaker_e ingredient_a),
(contains shaker_e ingredient_b),
(contains shaker_f ingredient_a),
(contains shaker_f ingredient_b),
(unshaked shaker_a),
(unshaked shaker_b),
(unshaked shaker_c),
(unshaked shaker_d)]

\end{verbatim}